\documentclass[letterpaper]{article} 
\usepackage{aaai24}  
\usepackage{times}  
\usepackage{helvet}  
\usepackage{courier}  
\usepackage[hyphens]{url}  
\usepackage{graphicx} 
\usepackage{natbib}  
\usepackage{caption} 
\usepackage{algorithm}
\usepackage{algorithmic}
\usepackage{bm}
\usepackage{amssymb}
\usepackage{latexsym}
\usepackage{bbding}
\usepackage{dsfont}
\usepackage{multirow}
\usepackage{color}

\newcommand{\tabincell}[2]{\begin{tabular}{@{}#1@{}}#2\end{tabular}}
\usepackage{newfloat}
\usepackage{listings}
\DeclareCaptionStyle{ruled}{labelfont=normalfont,labelsep=colon,strut=off} 
\floatstyle{ruled}
\newfloat{listing}{tb}{lst}{}
\floatname{listing}{Listing}

\title{Fewer Steps, Better Performance: Efficient Cross-Modal Clip Trimming for Video Moment Retrieval Using Language}
\author{
    Xiang Fang\textsuperscript{\rm 1}\equalcontrib, Daizong Liu\textsuperscript{\rm 2}\equalcontrib, Wanlong Fang\textsuperscript{\rm 3,1}\equalcontrib, Pan Zhou\textsuperscript{\rm 1}\thanks{Corresponding Author.}, Zichuan Xu\textsuperscript{\rm 4}, Wenzheng Xu\textsuperscript{\rm 5}, Junyang Chen\textsuperscript{\rm 6}, Renfu Li\textsuperscript{\rm 7} 
}
\affiliations{
    \textsuperscript{\rm 1}Hubei Engineering Research Center on Big Data Security, School of Cyber Science \\and Engineering, Huazhong University of Science of Technology Wuhan, China\\ 
    \textsuperscript{\rm 2}Peking University \\
    \textsuperscript{\rm 3}Henan University \\
    \textsuperscript{\rm 4}Dalian University of Technology \\
    \textsuperscript{\rm 5}Sichuan University \\
    \textsuperscript{\rm 6}Shenzhen University \\
    \textsuperscript{\rm 7}Huazhong University of Science and Technology

     xfang9508@gmail.com, dzliu@hust.edu.cn, wanlongfang@gmail.com, panzhou@hust.edu.cn, z.xu@dlut.edu.cn, wenzheng.xu@scu.edu.cn, junyangchen@szu.edu.cn, renfu.li@hust.edu.cn
}

\usepackage{bibentry}

\begin{document}

\maketitle

\begin{abstract}
Given an untrimmed video and a sentence query, video moment retrieval using language (VMR) aims to locate a target query-relevant moment. 
Since the untrimmed video is overlong, almost all existing VMR methods first sparsely down-sample each untrimmed video into multiple fixed-length video clips and then conduct multi-modal interactions with the query feature and expensive clip features for reasoning, which is infeasible for long real-world videos that span hours. Since the video is downsampled into  fixed-length clips,  some query-related frames may be filtered out, which will blur the specific boundary of the target moment, take the adjacent irrelevant frames as new boundaries, easily leading to cross-modal misalignment and introducing both boundary-bias and reasoning-bias.
To this end, in this paper, we propose an efficient approach, SpotVMR, to trim the query-relevant clip. Besides, our proposed SpotVMR can serve as plug-and-play module, which achieves efficiency for state-of-the-art VMR methods while maintaining good retrieval performance. 
Especially, we first design a novel clip search model that learns to identify promising video regions to search conditioned on the language query. Then, we introduce a  set of low-cost semantic indexing features to capture the context of objects and interactions that suggest where to search the query-relevant moment. Also, the distillation loss is utilized to  address the optimization issues arising from end-to-end joint training of the clip selector and VMR model.
Extensive experiments on three challenging datasets demonstrate its effectiveness.
\end{abstract}

\begin{figure}[t!]
\centering
\includegraphics[width=0.46\textwidth]{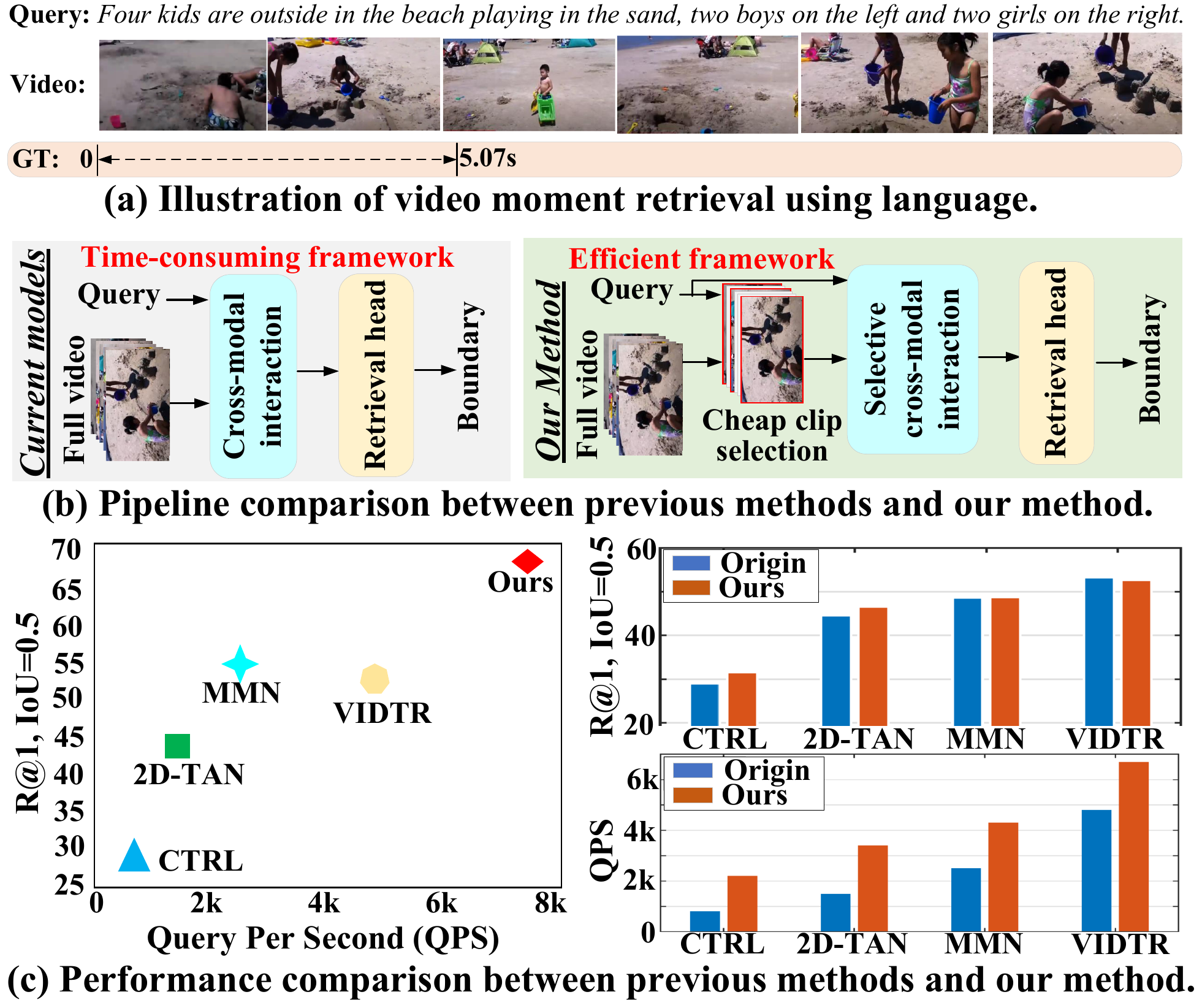}
\caption{ (a) Example of the VMR task, where  GT means the ground-truth boundary. (b) Pipeline comparison between previous  models and our  model. Previous models trim a long video into multiple fixed-length clips and perform costly processing of every clip. These processed clips are fed to the VMR model.
We propose an efficient  clip selection approach that adaptively
spots query-relevant clips quickly, and selectively processes these clips to serve as inputs to the VMR model. (c) Performance comparison with state-of-the-art VMR works  on  Charades-STA. Best viewed in color.}
\label{fig:intro}
\end{figure}

\section{Introduction}
\label{sec:intro}
As an emerging and challenging cross-modal task, video moment retrieval using language (VMR) \cite{anne2017localizing,gao2017tall} has drawn increasing attention in recent years due to its various applications, such as video understanding \cite{liu2023mllms,liu2020jointly,liu2021context,liu2023incomplete,liu2022memory,liu2021adaptive,liu2023transform,liu2023dicnet,liu2022exploring,liu2023information,liu2022unsupervised,fang2020v,fang2021animc,fang2021unbalanced} and temporal action localization \cite{zhang2020learning,fang2020double,liu2023exploring,wang2025taylor,fang2026towardsicml,kuai2026dynamic,wang2025point,fang2025your,zhang2025monoattack,fang2023hierarchical,liu2024towards,yang2025eood,fang2022multi,fang2026cogniVerse,lei2025exploring,fang2023you,wang2025dypolyseg,fang2025hierarchical,yan2026fit,fang2025adaptive,wang2026topadapter,cai2025imperceptible,fang2026slap,wang2026reasoning,fang2026immuno,wang2026biologically,fang2026disentangling,wang2025reducing,fang2026advancing,fang2026unveiling,wang2026from,liu2023conditional,liu2026attacking,fang2026rethinking,wang2025seeing,fang2026towards,fang2025multi,liu2024pandora,fang2024multi,fang2025turing,fang2024not,liu2023hypotheses,fang2024rethinking,liu2024unsupervised,fang2023annotations,xiong2024rethinking,fang2021unbalanced,wang2025prototype,zhang2025manipulating,fang2026align,tang2024reparameterization,fang2025adaptivetai,tang2025simplification,fang2021animc,cai2026towards,fang2020v,ji2023partial,ji2018semantic,ji2023online,ji2023binary,ji2023towards,ji2023mrtnet,ji2021vidvrd,ji2020context,ji2019human}.
As shown in Figure~\ref{fig:intro}(a), the VMR task targets locating  a  video moment that semantically corresponds to a given language query from a long untrimmed video.
Most of the video contents are query-irrelevant, where only a short video segment matches the query. It is substantially more challenging since a well-designed method needs to not only model the complex cross-modal interaction between videos and queries, but also capture complicated context information for cross-modal semantics alignment. Not only does it require recognizing objects and activities, but also identifying which visual content is sufficient to retrieve the accurate moment expressed in free-form natural language, accounting for the fact that the accurate moment may occupy only a tiny portion of the entire video.  Moreover, all this must be done in a scalable manner, given that a long untrimmed video (\textit{e.g.}, surveillance video and live video) will ultimately span hours, days or more.
In practice, VMR is an extremely challenging task because the desired model should (i) cover various  moment lengths in multiple scenarios; (ii) bridge the semantic gap between different modalities (video and query);
(iii) understand the semantic details of different modalities to extract modal-invariant features for optimal  retrieval.

Most previous VMR works \cite{zheng2023progressive,shen2023semantics,yang2022video,dong2022partially,dong2022dual,dong2022reading,dong2023dual,dong2023region,sun2023unified,ma2020fine,liu2018cross,liu2023filling,ge2019mac,zhang2019man,qu2023distantly,qu2021attend,wen2022survey,wen2020adaptive,wen2023deep} are under fully-supervised setting, where each frame is manually labeled as query-relevant or not.
Therefore, the main challenge in such a setting is how to align multi-modal features well to predict precise moment boundaries.
These fully-supervised approaches can be divided into two categories:
1) Top-down approaches \cite{anne2017localizing,chen2018temporally,zhang2019cross,zhang2020learning}: These methods integrate sentence information with each fine-grained video clip unit, and predict the similarity scores of candidate segment proposals by gradually merging the fusion feature sequence over time. The best proposal with the highest  score is selected as the predicted segment. 
2) Bottom-up approaches \cite{chenrethinking,mun2020local,zhang2020span}: These methods leverage the interaction between video and sentence to directly regress the start and end boundary frames of the target segment or predict boundary probabilities frame-wisely. The predicted segment is obtained through post-processing steps that group or aggregate all frame-wise predictions. 
Obviously, the frame-based annotation is very time-consuming, which will limit the applications of these methods. 
Although the above two types of works have achieved significant performances, they  still suffer from the redundant proposal generation/matching process (top-down) and complex post-processing steps (bottom-up) to refine the grounding results.

Although the above VMR methods have made exciting headway, they neglect the practical scaling issue: they extract expensive spatio-temporal visual features for densely sampled clips throughout the long video, which spends the most computational cost of a VMR model. Such a time-consuming approach becomes intractable as the video duration grows, especially for real-time applications like surveillance video and live video,
where the constrained on-board computation severely limits the applications of previous heavy-weight models. 
Fortunately, we can notice that ({i}) not all parts of the video are useful for reasoning about a given query, and ({ii}) there are high-level visual semantics about  objects and activities that could steer our attention toward where to retrieve. 
For example, given a query  ({``A person is hitting the sides of their bed with the palm of their hand."}), we can ignore video clips recorded in some irrelevant scenes other than the bedroom. Notably, these associations cannot be neatly enumerated, however, given the free-form nature of the queries. 
In the query ({``Four kids are outside in the beach playing in the sand, two boys on the left and two girls on the right"}),  the model reasoning will become  more complex since we need to reason about all persons in the beach (two boys and two girls), and identify their genders and position. Thus, we should learn query-conditioned priors that can use such high-level semantics to narrow down our  task.

In this paper, considering that the target moment is often located near key frames or key clips, we build upon these intuitions to propose a novel and effective approach to make a given VMR method more efficient and effective. The idea is to preview the video using cheap indexing features, intelligently select a small subset of \emph{query-relevant} clips, and only use these clips for the  target moment retrieval. This can cut down computational costs without sacrificing model performance.
To tackle this challenging setting with long-form  videos and language queries, we design a novel clip selection architecture, which introduces a cross-modal transformer to recursively preview the video and identify some query-relevant clips.  In the VMR task, the key visual features include three kinds: background feature to locate the place, appearance feature to detect the target instance, and the motion feature to recognize the activity mentioned in the query. Thus, 
we further design the above three kinds of  semantic-indexing features that capture video context about \textbf{b}ackground features, \textbf{a}ppearance features, and \textbf{m}otion features (\emph{BAM}).
With the high-level BAM visual features as the index, we can recursively update  the selected clips with the help of query features.
 Further, we design teacher-based distillation losses to optimize the cross-modal interaction.

To sum up, our main contributions are as follows:
\begin{itemize}
    \item In this paper, we target a novel and  efficient 
    clip selection approach for VMR,
     which first previews the video using cheap indexing features, then selects a small subset of query-relevant clips, and finally only uses these selected clips for the final  moment retrieval. 
    \item We propose an effective clip selection module by designing three high-level features (BAM features) as the semantic indexer. Then, an adaptive clip update strategy can update selected clips with a feature distillation loss as supervision during each iteration.
    
    \item  Our experiments on three popular yet challenging benchmarks demonstrate that our approach is more effective and efficient than state-of-the-art methods.
    
\end{itemize}

 \begin{figure*}[th!]
    \centering
    \includegraphics[width=\textwidth]{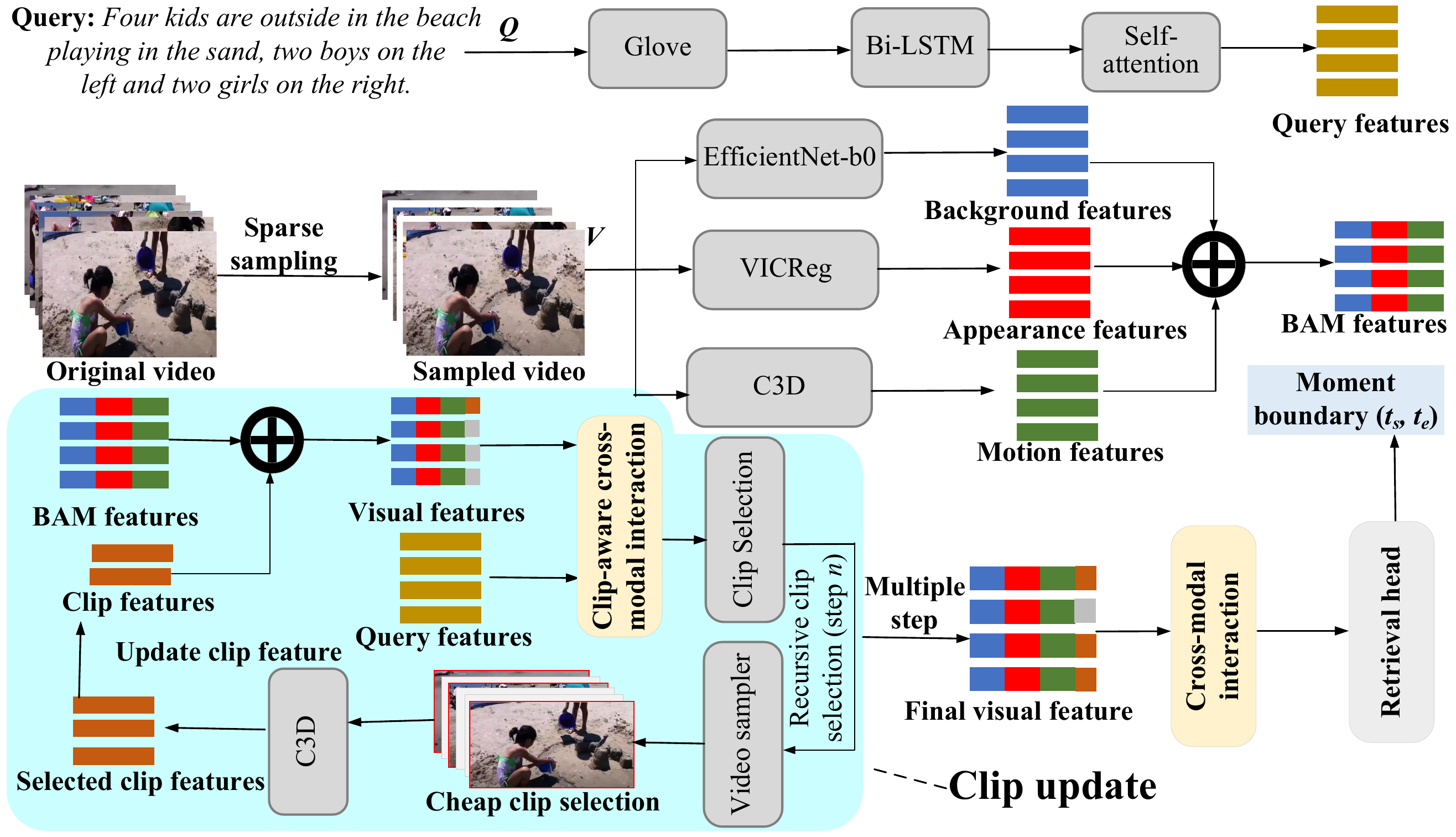}
    \caption{ Overview of our  model, which introduce a novel clip selection approach to search the core clip for VMR iteratively. We introduce a brand-new clip selection approach to search the core clip for VMR iteratively. Specially, we first extract the BAM semantic indexing features $s$ that capture background, appearance and motion context in the given untrimmed video by designing a cheap semantic indexer. Then, we feed the query into a textual encoder to obtain the word-aware query feature $q$. Moreover, we design a VMRSpotter to recursively select a subset of query-relevant clips. In the VMRSpotter, we first previews the video based on the semantic index $s$, then  alternate between selecting subsets of video clips $c_b$ for expensive feature extraction and previewing the video again with both the semantic index $s$ and the previously selected clip features $c_{b+1}$. By multiple steps of the VMRSpotter, we extract the cumulative selected clip features $c = c_{B+1}$ as the final clip features and the semantic index $s$ as the indexer for VMR. Besides, we design a cross-modal interaction model to jointly reason video feature $v$ and the query feature $q$ to obtain a cross-modal embedding $e$. Based on the cross-modal embeding, we introduce a retrieval head to predict the temporal extents of the moment boundary. To extract the precise feature, the cross-modal encoder inside the VMRSpotter module shares weights with the cross-modal encoder after  the VMRSpotter module. For the initial step (b=1), no clip is selected, i.e.,  $\bar{\bm{v}}_1 = \bm{0}$.}
    \label{fig:pipeline}
\end{figure*}

\section{Related Works}
Most existing VMR methods \cite{zhang2019cross,qu2020fine,liu2022skimming,liu2021progressively,liu2022reducing} can be divided into two categories:
1) Top-down methods \cite{gao2017tall,liu2022few,zhang2020learning,zeng2020dense,wang2022negative,wang-shi-2023-video,wang-etal-2023-balance,ijcai2021p156,li2023dual,li2023attribute,yang2023bicro}: They first pre-define multiple segment proposals and then align these proposals with the query for cross-modal semantic matching.
Finally, the best proposal with the highest similarity score is selected as the predicted segment. 
Although achieving decent results, these proposal-based methods severely rely on the quality of the segment proposals and are time-consuming.
2) Bottom-up methods \cite{zhang2020span,chenrethinking,mun2020local,tang2021frame,nan2021interventional,ji2023unbiased,ji2023higher,jian-wang-2023-invgc}: They directly regress the start and end boundary frames of the target segment or predict boundary probabilities frame-wisely. 
Compared with the proposal-based methods, proposal-free methods are more efficient. However, the above methods heavily rely on the datasets that require numerous manually labelled annotations for training.
In real-world applications, we always collect overlong videos. If we directly utilize the video for VMR, it will lead to much computational cost. Although some egocentric video search works \cite{jia2022egotaskqa,ramakrishnan2023spotem} are proposed to accelerate the video understanding process, they rely on the egocentric video and a language question, which is different from the inputs of VMR.
Thus, we present a brand-new  setting, called clip trimming VMR, with a merely light-weighted model rather than a large-weighted network.

\section{Proposed Method}
We propose  a light-weighted clip-selection approach that intelligently spots query-relevant clips for efficient VMR. Our  model consists of a novel clip selection architecture called ClipSpotter,  our BAM semantic indexing features to select latent query-relevant clips  for cross-modal interaction, and distillation loss to address optimization issues arising from jointly training ClipSpotter with the VMR task modules. ClipSpotter previews the video using our RIO features, which are obtained by selecting a single image from each clip and encoding them using efficient visual encoders~\cite{tan2019efficientnet}. Heavy clip features are then extracted from only the smaller subset of clips selected by ClipSpotter. 

\subsection{Overview}
\noindent \textbf{Problem statement.}
\label{sec:em_task}
Given an untrimmed video $\mathcal{V}$ with the frame number of $T$ and a sentence query $\mathcal{Q}$ composed of $N$ words, the task of Video Moment Retrieval Using Language (VMR) aims to precisely locate a temporal moment boundary $(\tau_s, \tau_e)$ in video $\mathcal{V}$, which starts at timestamp $\tau_s$ and ends at timestamp $\tau_e$, according to the semantics of query  $\mathcal{Q}$. 

\noindent \textbf{Pipeline.}
To tackle the light-weighted VMR, we propose a novel framework as shown in Figure~\ref{fig:pipeline}. For convenience, we define continuous 16 frames as a clip and each clip overlaps 8 frames with adjacent clips. We denote the number of clips as $C=T/8-1$.
First, a pretrained semantic indexer, which consists of one or more image encoders, is used to extract semantic index features.
After encoding the visual and textual features, we feed the multi-modal features into a RetrievalSpotter  module, which consists of a cross-modal interaction module and a selection policy to iteratively select the optimal clip.  Especially, in each iteration, we update the clip feature as the final clip feature.
After multiple steps, we can utilize the selective clip features for the VMR task.

\subsection{BAM Features for Semantic Indexing}
\label{sec:rio_feats}
To adaptively preview the video and select query-relevant clips, we target to learn the semantic indexer based on three high-level visual features: background feature, appearance feature and motion feature, termed BAM features. An intuitive idea is to employ the  efficient video recognition technology to extract the visual feature. However, previous efficient video recognition methods only capture the object-level appearance feature based on the pretrained ImageNet, and often ignore the background information and motion feature,
which is insufficient for our challenging VMR task. Since query-aware indexing in the VMR requires the background feature and motion information, we try to integrate the query feature and three high-level features (background, appearance and motion features) for previewing video.

For example, in a query \emph{``A woman and a man are sitting on the sidewalk playing music."}, the background-related text is \emph{``sidewalk"}, the appearance-related text is \emph{``a woman and a man"}, the motion-related text is \emph{``playing music"}. All three kinds of visual information are significant for the VMR task.
Hence, we design a set of low-cost semantic indexing features that capture context from the \textbf{b}ackground, object-level \textbf{a}ppearance, and  \textbf{m}otion, named \emph{BAM}. 

\noindent\textbf{Background features:} To effectively capture the background characteristics, we utilize a pre-trained EfficientNet-b0 image encoder \cite{tan2019efficientnet} as a background classifier. In the VMR task, a full video often contains less than three backgrounds, and the background change often corresponds to the start or end of an important video activity. 


\noindent\textbf{Appearance features:} In a video, the view is often  changed due to the object movement and the camera shift, which leads to the visual variance. To adaptively minimize the variance, we feed the frame-level video into a pretrained VICReg~\cite{bardes2022vicreg} network in a self-supervised way.
To extract appearance features accurately, we maintain diversity over each feature dimension, which can learn the object properties and closes the visual variance.


\noindent\textbf{Motion features:} For each clip, we first choose its start frame and end frame to extract the motion feature. Then, we feed the two frames into pretrained C3D network \cite{tran2015learning} to extract the clip-level motion features.


Overall, we sample one image within each video clip, extract each of the RIO features, and concatenate them to obtain the semantic indexing features $s= [s_1, s_2, \cdots, s_C] = {SemanticIndexer}(\mathcal{V})$.
These are image features extracted by sampling one image within each video clip, and they are inexpensive to compute.
These will serve as an initial preview of the video for intelligent clip selection.

\noindent \textbf{Query features.}
Similarly, given the query $Q$, we also follow  \cite{liu2022exploring,liu2023hypotheses} to utilize the Glove \cite{pennington2014glove} embedding to encode each word into a dense vector. We further employ the Bi-GRU \cite{chung2014empirical} layers to encode the word-level sequential information in the whole sentence. The final word-level feature can be denoted as $\bm{Q}=\{\bm{q}_j\}_{j=1}^{N} \in \mathbb{R}^{N \times D}$.
\small
\begin{equation}
    \label{eqn:textbackbone}
    q = [q_1, q_2, \cdots, q_N] = {QueryEncoder}(\mathcal{Q}).
\end{equation}\normalsize
Thus, by the semantic index $s$, query features $q$, and the video $\mathcal{V}$, we design a RetrievalSpotter module to recursively retrieve a final subset of query-relevant video clips $\mathcal{V}'$ as the moment candidates. Then, we feed all the frames of the selected clips $\mathcal{V}'$ into the same pre-trained C3D network to extract the expensive clip features. To avoid the unimportant computational cost, we set the features of those selected clips to zero in $v'= {VideoEncoder}(\mathcal{V}') \in \mathbb{R}^{C \times D}$.

\subsection{RetrievalSpotter Architecture}
\label{sec:selection_desc}
To adaptively select the query-relevant clips in the VMR task, we design a RetrievalSpotter network, which  first previews the entire video by an efficient semantic index, and then alternates between selected clips for expensive feature extraction. Finally, the RetrievalSpotter network repeats the above two processes until all the clips will be selected in the next recursive clip or the current recursive step reaches the maximum step. Thus, the semantic index $s$ is computed once before step 1 and kept fixed.


Specifically, we denote the recursive step as $b \in [1, \cdots, B]$, where $B$ is the  maximum step. $v'_b \in \mathbb{R}^{C \times D}$ denote the clip features for selected from steps $1$ to $b-1$, where $v'_1=[0]_{C \times D}$ is an all-zero matrix. For any clip feature $c_b$, we concatenate it with  $s$ along the feature dimension to fuse them as $s_b \in \mathbb{R}^{C \times 2D}$. To interact the visual and textual features for further reasoning, we perform the clip-aware cross-modal fusion as: $c'_b = {CrossModalFusion}(s_b, q)$,
where $c'_b$ is the fused cross-modal feature. 

By treating the fused feature as  guidance, we utilize a two-layered MLP as  the clip selection module to show if a clip feature should be computed or not. Especially, the clip selection module will output a binary value (\textit{i.e.}, thresholded probabilities) for each clip:
 \small
\begin{equation}
    \label{eqn:binarypred}
    p_{b+1} = {ClipSelection}(c'_b) \in \{0, 1\}^C,
\end{equation}\normalsize
where $p_{b+1}$ is the binary value, where if $p_{b+1}=1$, the clip will be selected and vice versa.
Finally, we feed these selected clips into the video encoder (C3D) to obtain the expensive visual clip features. We repeat the  selection process for steps, and utilize a cumulative set of clip features ($f_v = v'_{B+1}$) to predict the moment boundary.


To enhance the visual features with query-specific information, we perform  cross-modal interaction by  the concatenated clip and semantic index features $s\oplus v$ and the query features $q$.
\small
\begin{equation}
    \label{eqn:crossmodal}
    c = {CrossModalInteraction}(s\oplus v, q) \in \mathbb{R}^{C \times D_h}.
\end{equation}\normalsize
Finally, we design a retrieval module to  predict the temporal extent of the boundary $\hat{\mathcal{B}}$: $\hat{\mathcal{B}} = [\tau_s, \tau_e] = {Retrieval}(c)$.

Thanks to the above processes, we can enhance the efficiency of the state-of-the-art VMR methods. Specifically, the VMR works (\textit{e.g.}, VSLNet~\cite{zhang2020span} and MMN) feed the entire video into the visual encoder to extract the clip features,  utilize a query encoder to extract query features, use a cross-model interaction module to fuse visual and textual feature,  and finally employ a retrieval head module to predict the moment boundary. 
Our proposed SpotVMR can significantly save the computational cost by first previewing the video cheaply using the semantic indexer and then recursively selecting a subset of query-relevant clips by RetrievalSpotter.
By iteratively selecting a subset of clips for expensive feature extraction, our SpotVMR can modulate the video inputs to these state-of-the-art VMR models. Although different models utilize various cross-modal interaction modules and retrieval head modules, our efficient clip selection strategy still works.



\subsection{Model Optimization}
\label{sec:optim}
After selecting the query-relevant clips, we jointly optimize the cross-modal interaction and the retrieval head modules end-to-end to improve the retrieval performance. During training, we keep the video, semantic index, and text encoders frozen. Besides, multiple loss functions are introduced: a VMR task loss $\mathcal{L}_{vmr}$, a clip selection loss $\mathcal{L}_{sel}$,  and a novel feature distillation loss $\mathcal{L}_{ftd}$.


\noindent\textbf{Clip selection loss.} 
To select accurate clips to avoid  under-sampling and over-sampling, we introduce a clip selection loss: $L_{sel} = (\gamma-\mathbb{E}_{(v,q) \sim D_t} [\frac{1}{L}\sum_{l=1}^L {b}_{joint}^l] )^2$, where $D_{t}$ denotes the training dataset and $\bar{\bm{b}}_{joint} = \sum_{n=1}^{N+1} p_b$ is the overall binary selections after $N$ steps. By predefining the hyperparameter $\gamma$, the clip selection loss can limit the fraction of selected clips in expectation.  To regularize the per-step clip selection $p_b^l$, we encourage our proposed model to select $(\gamma L/{B})$ clips in each step.
Experimental results show that the above simple regularization can significantly improve training stability. Since  our proposed RetrievalSpotter predicts binary  values during clip selection, it is not differentiable for gradient-based optimization. Therefore, we introduce the Gumbel-Softmax trick to reparameterize argmax sampling using a softmax relaxation during training~\cite{hazan2012partition,wu2019liteeval}. 

\noindent\textbf{VMR task losses.}
We denote the video and query features as $v$ and $q$. By fusing $v$ and $q$, we utilize the cross-modal interaction module to obtain the cross-modal representation as $c = {CrossModalInteraction}(v, q) \in \mathbb{R}^{C\times D}$.
Especially, the module includes a transformer encoder module, which utilizes the self-attention process to update the video feature $v$ and query feature $q$ independently. Then, we utilize the context-query attention mechanism for enhancing the video features with the help of the query features~\cite{zhang2020span,seo2016bidirectional}. Then, we introduce a 1D convolutional layer to compute the probability that a clip lies within a temporal neighborhood of the target moment: $\hat{\mathcal{S}}_h = \sigma({Conv1D}(c)) \in \mathbb{R}^{C\times 1}$,
where $\sigma$ is the sigmoid  function, and $\hat{\mathcal{S}}_h$ is used to update the cross-modal features: $c = \hat{\mathcal{S}}_h \cdot c \in \mathbb{R}^{C \times D_h}$.
To infer the moment boundary, we introduce a retrieval module, including a transformer encoder for performing self-attention and an MLP layer to predict the log probabilities:
\small
\begin{equation}
    \label{suppeq:span}
    \hat{\tau}_s, \hat{\tau}_e = {RetrievalPrediction}(c),
\end{equation}\normalsize
where $\hat{\tau}_s, \hat{\tau}_e \in \mathbb{R}^{C\times 1}$ are log-probabilities per feature location, ``RetrievalPrediction'' means the retrieval module. We use the following loss to supervise the boundary predictions:
\small
\begin{equation}
   \mathcal{L}_{{boundary}} = L_{CE}(\hat{p}_s, p^{*}_s) + L_{CE}(\hat{p}_e, p^{*}_e),
\end{equation}\normalsize
where $L_{CE}$ denotes the cross-entropy loss, and $p^{*}_s, p^{*}_e$ are the ground-truth boundary of the target moment. We supervise the query-aware visual enhancement by the following loss:
\small
\begin{equation}
   \mathcal{L}_{{qav}} = f_{CE}(\hat{S}_h, S^{*}_h ),
\end{equation}\normalsize
where $S^{*}_h$ denotes the ground-truth enhancement score, which covers an extended temporal window around the ground-truth moment boundary. By jointing the above loss, we can obtain the overall VMR loss as follows:
\small
\begin{equation}
   \mathcal{L}_{vmr} = \mathcal{L}_{{boundary}} + \mathcal{L}_{{qav}}.
\end{equation}\normalsize


\noindent\textbf{Distillation loss:} 
To further fine-tune the moment boundary, we design a two-stage training strategy based on the knowledge distillation approach. First, we train a teacher VMR model without the RetrievalSpotter module as the distillation supervision. Then, we utilize a student VMR module with the RetrievalSpotter module for joint optimization.

Given a video-query pair $(V, Q)$ and its ground-truth moment boundary $B$, we denote the cross-modal interaction outputs  for the teacher and student VMR modules as $c_{teacher}$ and $c_{student}$, respectively. 
Different from the student module, we feed all the video features into the teacher module. To match the cross-modal features between the student module and the teacher module, we design the following {feature distillation loss}: $\mathcal{L}_{ftd} = ||{StopGrad}({c_b}_{teacher}) - {c_b}_{student}||_2$,
where $||\cdot||_2$ is the L-2 loss,
the gradient is not propagated to the frozen teacher.

Thus, our final loss is:
\begin{equation}
    \label{eqn:loss_distill}
    \mathcal{L}_{final} = \alpha \mathcal{L}_{vmr} + \beta \mathcal{L}_{sel} +\gamma \mathcal{L}_{ftd},
\end{equation}
where  $\alpha$, $\beta$ and $\gamma$ are  hyperparameters to balance the weights of different losses. By jointly training these losses, we encourage the model to improve VMR performance while limiting the budget of clips selected.

\section{Experiments}
\label{sec:exp_setup}

\begin{table}[t!]
\small
\begin{center}
\scalebox{1.0}{
\setlength{\tabcolsep}{1.4mm}{
\begin{tabular}{l|c|cccc}
    \hline
\multicolumn{6}{c}{Performance comparisons on  ActivityNet Captions}\\\hline
    \multirow{2}*{Method}  & \multirow{2}*{Type} & R@1, & R@1, & R@5, & R@5, \\ 
    ~ & ~ & IoU=0.5 & IoU=0.7 & IoU=0.5 & IoU=0.7 \\ \hline 
    CTRL & $\downarrow$ & 29.01 & 10.34 & 59.17 & 37.54 \\
    SCDM & $\downarrow$ & 36.75 & 19.86 & 64.99 & 41.53 \\
    CMIN &$\downarrow$ & 43.40 & 23.88 & 67.95 & 50.73 \\
    2D-TAN & $\downarrow$ & 44.51 & 26.54 & 77.13 & 61.96 \\
    DRN & $\downarrow$ & 45.45 & 24.36 & 77.97 & 50.30 \\
    MMN & $\downarrow$ & 48.59& 29.26& 79.50& 64.76\\\hline 
    GDP & $\uparrow$ & 39.27 & - & - & - \\
    LGI & $\uparrow$ & 41.51 & 23.07 & - & - \\ 
    VSLNet & $\uparrow$ & 43.22 & 26.16 & - & - \\
    IVG-DCL & $\uparrow$ & 43.84 & 27.10 & - & - \\ \hline 
    \textbf{Ours} & $\updownarrow$ & \textbf{52.83} & \textbf{32.76} & \textbf{84.37} & \textbf{68.95} \\ \hline
        \hline
\multicolumn{6}{c}{Performance comparisons on  Charades-STA}\\\hline
    \multirow{2}*{Method}  & \multirow{2}*{Type} & R@1, & R@1, & R@5, & R@5, \\ 
    ~ & ~ & IoU=0.5 & IoU=0.7 & IoU=0.5 & IoU=0.7 \\ \hline 
    CTRL & $\downarrow$ & 23.63 & 8.89 & 58.92 & 29.57  \\
    SCDM & $\downarrow$ & 54.44 & 33.43 & 74.43 & 58.08 \\
    2D-TAN & $\downarrow$ & 39.81 & 23.25 & 79.33 & 51.15  \\
    DRN & $\downarrow$ & 53.09 & 31.75 & 89.06 & 60.05  \\
    MMN& $\downarrow$ & 47.31& 27.28& 83.74& 58.41\\ \hline 
    GDP & $\uparrow$ & 39.47 & 18.49 & - & - \\
    VSLNet & $\uparrow$ & 47.31 & 30.19 & - & - \\
    IVG-DCL & $\uparrow$ & 50.24 & 32.88 & - & -  \\
    ACRM & $\uparrow$ & 57.53 & 38.33 & - & - \\
    \hline 
    \textbf{Ours} & $\updownarrow$ & \textbf{68.82} & \textbf{47.39} & \textbf{97.01} & \textbf{75.38}  \\ \hline
        \hline
\multicolumn{6}{c}{Performance comparisons on   TACoS }\\\hline
        \multirow{2}*{Method}  & \multirow{2}*{Type} & R@1, & R@1, & R@5, & R@5, \\ 
    ~ & ~ & IoU=0.3 & IoU=0.5 & IoU=0.3 & IoU=0.5 \\ \hline \hline
CTRL & $\downarrow$  & 18.32 & 13.30 & 36.69 & 25.42 \\
    SCDM & $\downarrow$  & 26.11 & 21.17 & 40.16 & 32.18 \\
    CMIN & $\downarrow$  & 24.64 & 18.05 & 38.46 & 27.02 \\
    2D-TAN & $\downarrow$  & 37.29 & 25.32 & 57.81 & 45.03 \\
    DRN & $\downarrow$  & - & 23.17 & - & 33.36 \\
    MMN & $\downarrow$  & 39.24& 26.17& 62.03& 47.39\\\hline \hline
    GDP & $\uparrow$ & 24.14 & - & - & - \\
    VSLNet & $\uparrow$ & 29.61 & 24.27 & - & - \\
    IVG-DCL & $\uparrow$ & 38.84 & 29.07 & - & - \\
    ACRM & $\uparrow$ & 38.79 & 26.94 & - & - \\
    \hline \hline
\textbf{Ours} & $\updownarrow$ & \textbf{48.72} & \textbf{38.94} & \textbf{67.03} & \textbf{56.38} \\
\hline
\end{tabular}}}
\end{center}
\caption{ Performance comparisons on three challenging datasets (top: ActivityNet Captions, middle: Charades-STA, bottom:  TACoS), where $\downarrow$ means the top-down setting; $\uparrow$ means the bottom-up setting, and $\updownarrow$ means our posed setting.}
\label{tab:three_dataset}
\end{table}

\noindent \textbf{Dataset.}
For  fair comparison with existing VMR works, we utilize the same datasets for evaluation:
ActivityNet Caption \cite{caba2015activitynet}, TACoS \cite{regneri2013grounding}, and Charades-STA \cite{sigurdsson2016hollywood}.
Specifically, ActivityNet Caption
contains 20000 untrimmed videos with 100000 descriptions from YouTube. Following the public split, we use 37417, 17505, and 17031 sentence-video pairs for training, validation, and testing. 
TACoS contains 127 videos collected from cooking scenarios. We also follow the public split, which includes 10146, 4589, 4083 query-segment pairs for training, validation and testing. As for Charades-STA, there are 12408 and 3720 moment-query pairs in the training and testing sets, respectively.



\noindent \textbf{Evaluation metrics.} 
Following
\cite{gao2017tall,zhang2020span},
we adopt ``R@n, IoU=m'' as the evaluation metrics. The ``R@n, IoU=m'' denotes the percentage of language queries having at least one result whose Intersection over Union (IoU) with ground truth is larger than m in top-n retrieved segment. In our experiments, we use $n \in \{1,5\}$ for all datasets, $m \in \{0.5,0.7\}$ for ActivityNet Captions and Charades-STA, $m \in \{0.3,0.5\}$ for TACoS.

\noindent \textbf{Implementation details.}
To encode each video, we 
define continuous 16 frames as a clip and each clip overlaps 8 frames with adjacent clips. Following previous works \cite{zhang2020learning,wang2022negative}, 
we employ the Glove model \cite{pennington2014glove} to embed each word to 300 dimension features. We train our whole model for 100 epochs with an early stopping strategy. Parameter optimization is performed by Adam optimizer with a learning rate of 0.0005, and a linear decay rate of 1.0. All the experiments are implemented by PyTorch. For the hyper-parameters, we set  $\alpha=0.4$, $\beta=0.8$, and $\gamma=0.6$.

\begin{table}[t!]
\small
    \centering
      \scalebox{1}{
    \setlength{\tabcolsep}{2.1mm}{
    \begin{tabular}{c|cccc|c|ccccccccccccc}
    \hline
    \multirow{2}*{Model} & \multicolumn{4}{|c|}{$T_{ext}$} &\multirow{2}*{$T_{exe}$}& \multirow{2}*{$T_{total}$} \\ \cline{2-5}  
    ~ & B&A&M& Other&~&~\\ \hline
        CTRL &-&-&-&18.51&187.52&206.03\\
        RaNet&-&-&-&18.51&208.40&226.91\\
    2D-TAN&-&-&-&18.51&216.87&235.38\\
        MIGCN &-&-&-&18.51&253.94&271.89\\
        MMN &-&-&-&18.51&289.31&207.82\\
        DRN &-&-&-&18.51&294.70&313.21\\\hline
    \textbf{Ours}&\textbf{1.02}&\textbf{1.34}&\textbf{2.57}&-&\textbf{32.75} &\textbf{37.68}\\\hline
    \end{tabular}}}
    \caption{Efficiency comparison (time complexity (s) of 100 videos) on  ActivityNet Captions. The total time $T_{total}$ comprises the measurement time of extracting the corresponding features ($T_{ext}$), and executing the network models ($T_{exe}$), where ``Other'' means the feature encoder (\textit{e.g.,} C3D/I3D).}
    \label{tab:efficiency}
\end{table}

\subsection{Comparison with State-of-the-Arts}
For performance evaluation, we compare several state-of-the-art open-source VMR methods that are grouped into two categories: 
1) Top-down ($\downarrow$): CTRL \cite{gao2017tall},  SCDM \cite{yuan2019semantic}, CMIN \cite{zhang2019cross}, 2D-TAN \cite{zhang2020learning}, DRN \cite{zeng2020dense}, MMN \cite{wang2022negative}; 
2)  Bottom-up ($\uparrow$):  GDP \cite{chenrethinking}, LGI \cite{mun2020local}, VSLNet \cite{zhang2020span}, IVG-DCL \cite{nan2021interventional}, ACRM \cite{tang2021frame}.
The best results are \textbf{bold}. As shown in
Table \ref{tab:three_dataset}, our model beats all compared methods by a large margin, which illustrates the effectiveness of our model.

\begin{table*}[t!]
    \centering
    \small
    \setlength{\tabcolsep}{0.9mm}{
    \begin{tabular}{c|cccc|cccc|cccc}
    \hline
    \multirow{3}*{Method}  & \multicolumn{4}{c|}{ActivityNet Captions} & \multicolumn{4}{c|}{Charades-STA} & \multicolumn{4}{c}{TACoS} \\ \cline{2-5} \cline{6-9} \cline{10-13}
    ~ & R@1, & R@1, & R@5, & R@5, & R@1, & R@1, & R@5, & R@5, & R@1, & R@1, & R@5, & R@5, \\
    ~ & IoU=0.5 & IoU=0.7 & IoU=0.5 & IoU=0.7& IoU=0.5 & IoU=0.7 & IoU=0.5 & IoU=0.7  & IoU=0.3 & IoU=0.5 & IoU=0.3 & IoU=0.5 \\ \hline
    w/o $\mathcal{L}_{qav}$ &48.72&29.53&82.53&67.30&68.01&45.32&94.27&73.52&46.80&37.85&65.29&53.71 \\
    w/o $\mathcal{L}_{sel}$ & 50.66&30.80& 81.32&66.97&67.35&44.86&95.41&74.25&47.13&37.92&66.17&54.82\\
    w/o $\mathcal{L}_{ftd}$ & 51.39& 31.95&84.03&68.54&67.90&46.81& 96.57&75.32&48.08&38.75&66.88&54.18 \\
    \hline
    Full model & \textbf{52.83} & \textbf{32.76} & \textbf{84.37} & \textbf{68.95}& \textbf{68.82} & \textbf{47.39} & \textbf{97.01} & \textbf{75.38} &\textbf{48.72} & \textbf{38.94} & \textbf{67.03} & \textbf{56.38} \\\hline
    \end{tabular}}
    \caption{ Main ablation study on all the datasets, where we remove each key individual module to investigate its contribution.}
    \label{tab:ablation1}
\end{table*}
\begin{table*}[th!]
\centering
\small
\setlength{\tabcolsep}{0.9mm}{
\begin{tabular}{c|c|cccc|cccc|ccccccccccc}
\hline
\multirow{3}*{Model}&\multirow{3}*{Setting} & \multicolumn{4}{c|}{ActivityNet Captions}& \multicolumn{4}{c|}{Charades-STA}  & \multicolumn{4}{c}{TACoS}  \\\cline{3-14}
~&~ & R@1  & R@1 & R@5  & R@5& R@1  & R@1 & R@5  & R@5& R@1  & R@1 & R@5 & R@5 \\
~&~   & IoU=0.5 & IoU=0.7 & IoU=0.5 & IoU=0.7& IoU=0.5 & IoU=0.7 & IoU=0.5 & IoU=0.7  & IoU=0.3 & IoU=0.5 & IoU=0.3 & IoU=0.5 \\\hline
\multirow{2}*{2D-TAN} &Origin&44.51 & 26.54 & 77.13 & 61.96& 39.81 & 23.25 & 79.33 & 51.15& 37.29 & 25.32 & 57.81 & 45.03\\
~&\textbf{Ours} & \textbf{46.21}&\textbf{27.43}& \textbf{78.64}& \textbf{63.58}& \textbf{42.95}& \textbf{25.10}& \textbf{81.72}& \textbf{53.27}& \textbf{38.49}& \textbf{26.15}& \textbf{59.22}& \textbf{45.83}\\\hline
\multirow{2}*{MMN}&Origin& 48.59& 29.26& 79.50& 64.76& 47.31& 27.28& 83.74& 58.41&39.24& 26.17& 62.03& 47.39\\
~&\textbf{Ours}  &\textbf{49.37}& \textbf{30.52}& \textbf{80.26}& \textbf{65.31}& \textbf{49.12}& \textbf{28.54}& \textbf{85.43}& \textbf{59.72}& \textbf{41.17}& \textbf{27.43}& \textbf{63.92}& \textbf{49.30}  \\\hline
\end{tabular}}
\caption{ Our proposed method serves as a plug-and-play module for state-of-the-art models on different datasets.}
\label{tab:plug}
\end{table*}

\noindent \textbf{Efficiency comparison.} 
As shown in Table \ref{tab:efficiency}, we conduct the efficiency comparison on  ActivityNet Captions with some state-of-the-art open-source methods. Our model is more  efficient than compared methods by a large margin.

\begin{table}[t]
\small
\centering
\begin{tabular}{@{}ccc|cccccc@{}}
\hline
\multirow{2}*{B}& \multirow{2}*{A}& \multirow{2}*{M}& R@1,   & R@1,   & R@5, & R@5, \\
&&&IoU=0.5&IoU=0.7&IoU=0.5&IoU=0.7\\\hline
\CheckmarkBold & \XSolidBrush & \CheckmarkBold &50.24&30.62&82.49&68.43 \\
\CheckmarkBold & \CheckmarkBold & \XSolidBrush & 51.35&31.84&82.88&68.27\\
\XSolidBrush & \CheckmarkBold & \CheckmarkBold & 51.98& 32.25& 83.71&68.60\\\hline
\CheckmarkBold &\CheckmarkBold &\CheckmarkBold &\textbf{52.83} & \textbf{32.76} & \textbf{84.37} & \textbf{68.95}\\
\hline
\end{tabular}
\caption{Effect of semantic index on  ActivityNet Captions. }
\label{tab:ablation_feats}
\end{table}

\subsection{Ablation Study}
\noindent \textbf{Main ablation study.}
To demonstrate the effectiveness of each component in our model, we conduct ablation studies regarding the components. The corresponding experimental results are reported in Table \ref{tab:ablation1}. Obviously, we can find that both two modules contribute a lot to the final performances, showing that  each module is effective for the VMR task. 

\begin{table}[t!]
\small
\scalebox{1.0}{
\setlength{\tabcolsep}{1.2mm}{
\begin{tabular}{cc|cccccc}
\hline
\multirow{2}*{Module} &\multirow{2}*{Changes} & R@1 & R@1 & R@5 & R@5\\
&~ & IoU=0.5 & IoU=0.7 & IoU=0.5 & IoU=0.7\\
\hline
\multirow{3}*{\tabincell{c}{Recursive  \\ steps}} &3&51.92& 31.40&83.55&67.38\\
~&\textbf{5}& 52.83& \textbf{32.76}&\textbf{84.37}&\textbf{68.95} \\
~&7&\textbf{53.04}&31.95&83.52&68.24 \\ \hline
\end{tabular}}}
\caption{ Effect of recursive step on   ActivityNet Captions.}
\label{tab:steps}
\end{table}
\begin{figure}[t!]
\centering
\includegraphics[width=0.23\textwidth]{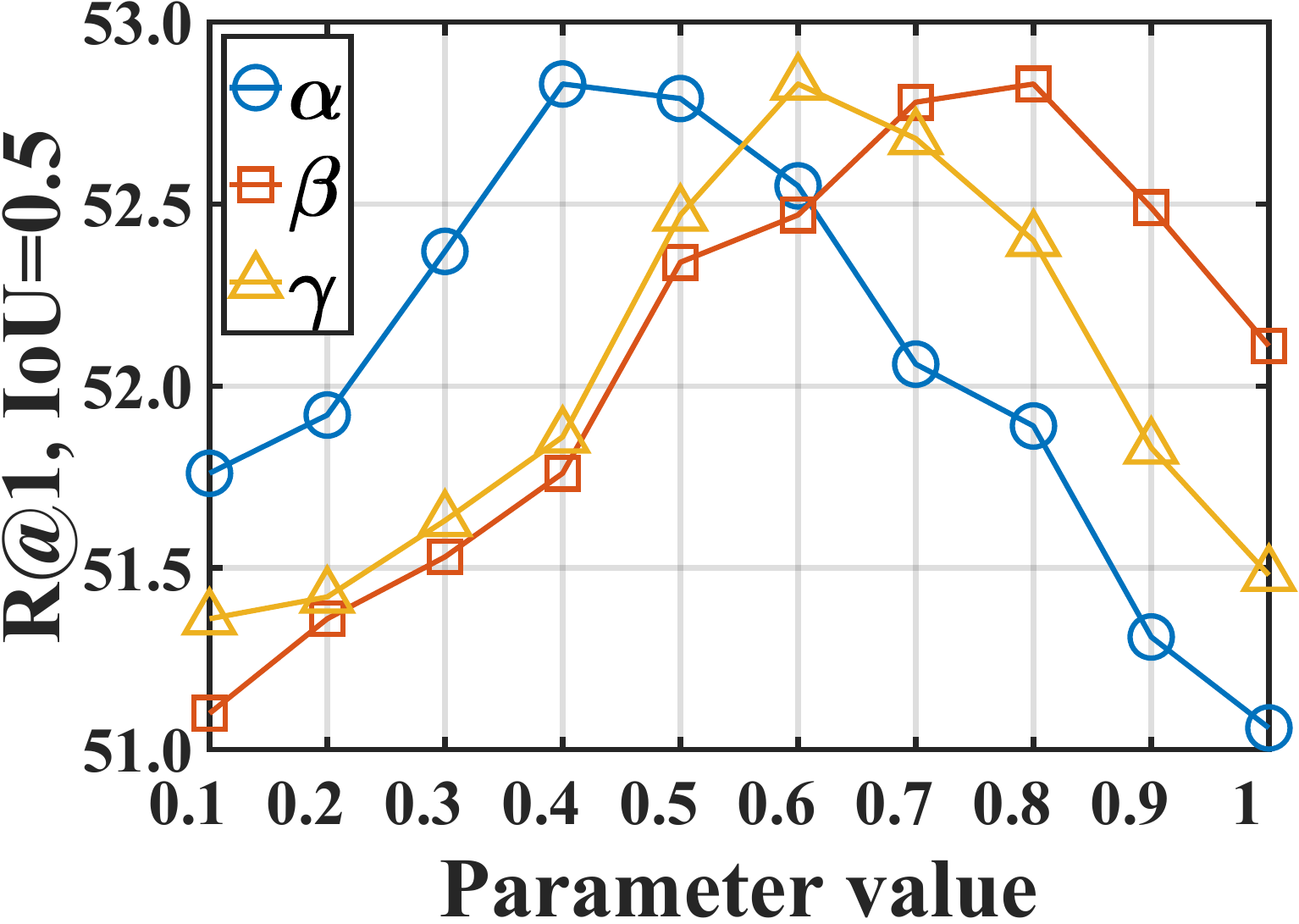}
\hspace{-0.1in}
\includegraphics[width=0.23\textwidth]{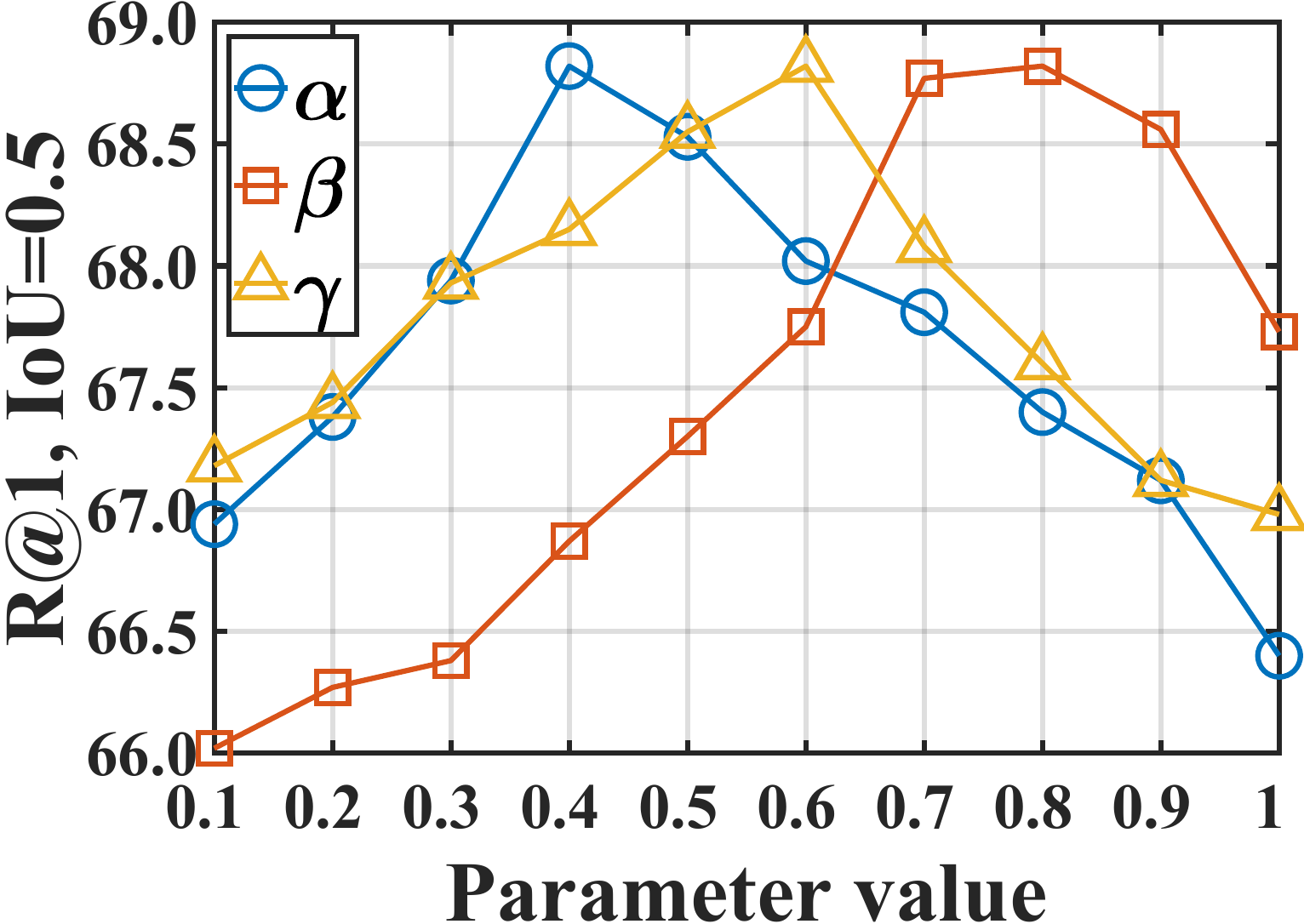}
\caption{ Analysis on the parameters ($\alpha,\beta,\gamma$) on  ActivityNet Captions (left) and Charades-STA (right).}
\label{fig:canshu}
\end{figure}

\noindent \textbf{Plug-and-play.} To further compare with current methods, we serve our method as a plug-and-play module for state-of-the-art models (2D-TAN and MMN). As shown in Table \ref{tab:plug}, our method can significantly improve their performance, which shows the effectiveness of our method.

\noindent \textbf{Effect of the BAM feature.} To analyze the contribution of different high-level features, we conduct the ablation study in Table \ref{tab:ablation_feats}.  Background(B), appearance(A) and motion(M) features can significantly improve the performance. The improvement shows the effectiveness of our designed features. During the clip selection, we set the maximum recursive
step as $B_{\max}$, we analyze the effect of different $B_{\max}$ on Table \ref{tab:steps}. When $B_{\max}=5$, we can obtain the best performance. 



\begin{figure}[t!]
\centering
\includegraphics[width=0.46\textwidth]{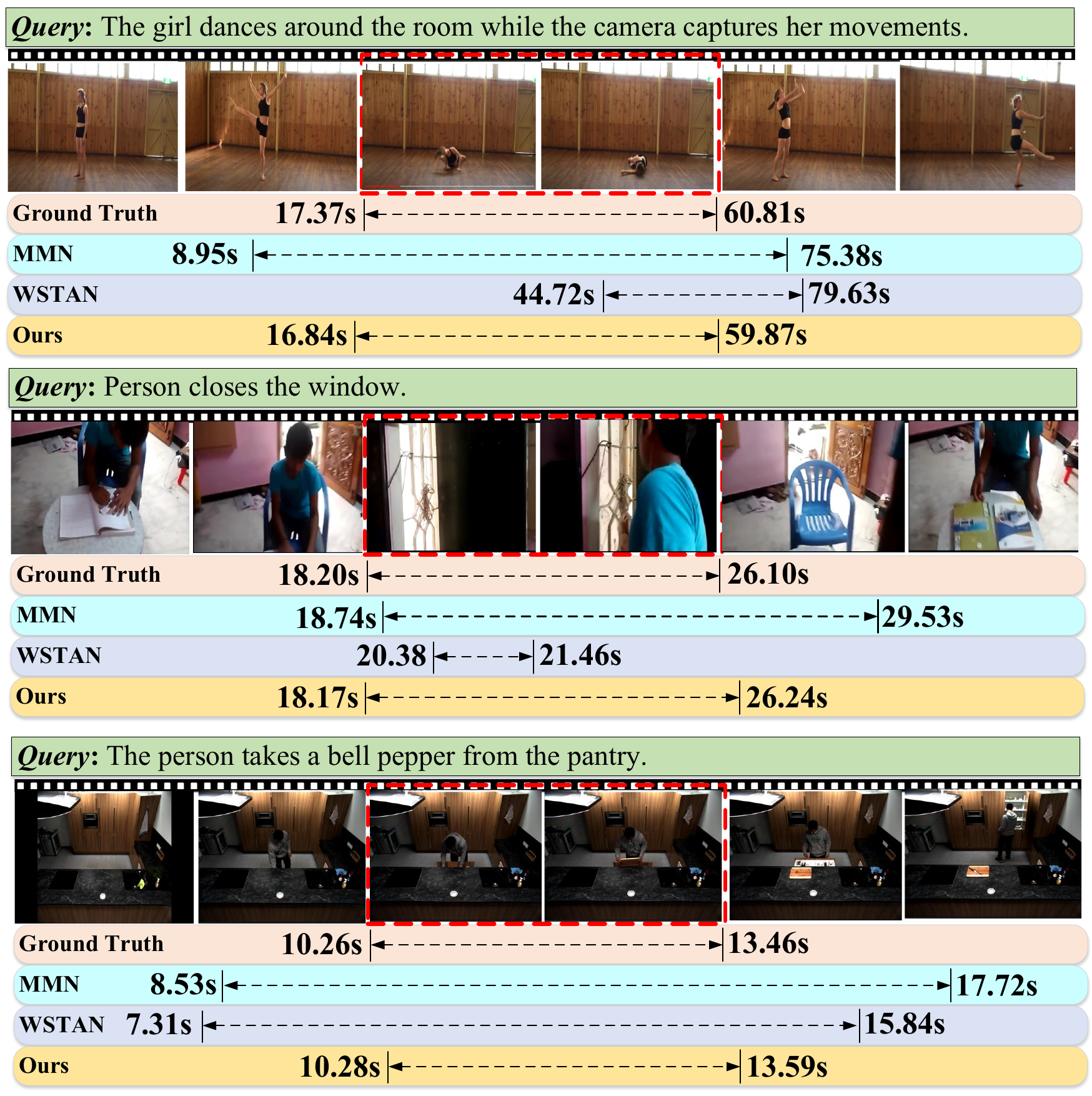}
\caption{ Qualitative results on three datasets (top: ActivityNet Captions, middle: Charades-STA, bottom: TACoS).}
\label{fig:result}
\end{figure}

\noindent \textbf{Analysis on parameters.}
We conduct the ablation studies on the parameters $\alpha,\beta,\gamma$ in Figure~\ref{fig:canshu}. 
Specifically, we change  one  parameter with fixing the others.
We obtain the best performance when  $\alpha=0.4,\beta=0.6,\gamma=0.8$.

\subsection{Qualitative results}
We provide the retrieval visualizations on three datasets in Figure~\ref{fig:result}. Our method can retrieve  more precise moment boundaries than previous state-of-the-art methods (MMN \cite{wang2022negative} and WSTAN \cite{wang2021weakly}).


\section{Conclusion}
In this paper, we propose a novel and efficient video moment retrieval setting, which first previews the whole video by a semantic indexer, and then retrieves the target moment boundary by a distillation loss.  Experiments on three challenging datasets 
show the effectiveness of our  method.

\section{Acknowledgments}
This work is supported by National Natural Science Foundation of China (NSFC) under grant no. 61972448 and  no. 62272328.

{\small \bibliography{aaai24}}

\end{document}